\documentclass{article}

    \usepackage[final]{neurips_2024}

\usepackage[utf8]{inputenc} 
\usepackage[T1]{fontenc}    
\usepackage{hyperref}       
\usepackage{url}            
\usepackage{booktabs}       
\usepackage{amsfonts}       
\usepackage{nicefrac}       
\usepackage{microtype}      
\usepackage{xcolor}         
\usepackage{graphicx} 
\usepackage[
backend=biber,
style=phys,
]{biblatex}
\usepackage{subcaption}
\usepackage{amsmath}

\title{Inferring Stability Properties of Chaotic Systems on Autoencoders' Latent Spaces}

\author{%
 Elise Özalp \\
  Department of Aeronautics\\
  Imperial College London\\
  \texttt{elise.ozalp@imperial.ac.uk} \\
  \And
  Luca Magri \\
  Department of Aeronautics\\
  Imperial College London\\
  The Alan Turing Institute\\
  \texttt{l.magri@imperial.ac.uk} \\
}

\begin{document}

\maketitle

\begin{abstract}

%
The data-driven learning of solutions of partial differential equations can be based on a divide-and-conquer strategy. First, the high dimensional data is compressed to a latent space with an autoencoder; and, second, the temporal dynamics are inferred on the latent space with a form of recurrent neural network. In chaotic systems and turbulence, convolutional autoencoders and echo state networks (CAE-ESN) successfully forecast the dynamics, but little is known about whether the stability properties can also be inferred. We show that the CAE-ESN model infers the invariant stability properties and the geometry of the tangent space in the low-dimensional manifold (i.e. the latent space) through Lyapunov exponents and covariant Lyapunov vectors. This work opens up new opportunities for inferring the stability of high-dimensional chaotic systems in latent spaces.
\end{abstract}

\begin{figure}[h]
    \centering
    \includegraphics[width=0.9\linewidth]{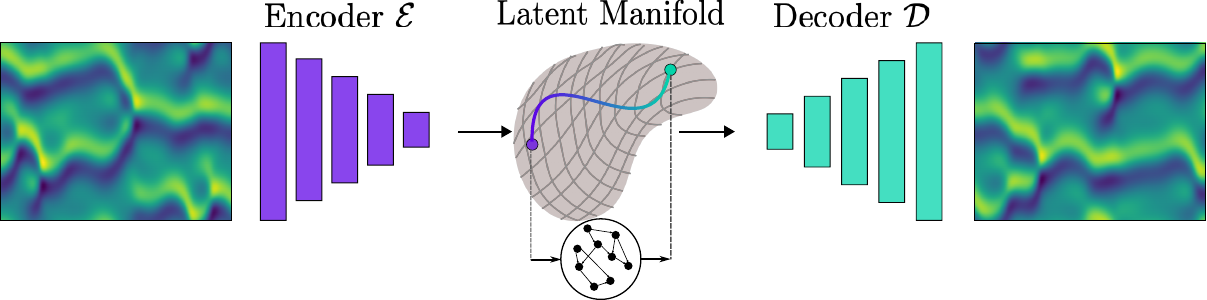}
    \caption{Illustration of the CAE-ESN with the Kuramoto-Sivashinsky data as an example.}
    \label{fig:ks_autoencoder_potatoe_esn}
\end{figure}
\section{Introduction}

Convolutional autoencoder recurrent neural networks have become a promising approach for forecasting chaotic and turbulent systems \cite{linot2022data, vlachas2022multiscale, racca2023predicting, fukami2020convolutional}. This framework addresses the challenge of modelling high-dimensional chaotic systems by compressing the data into a nonlinear latent representation through a convolutional autoencoder (CAE) and a recurrent neural network to propagate the lower-dimensional temporal dynamics on the latent manifold. This addresses a challenge in chaotic partial differential equations (PDEs) and turbulent systems: although the state might live in a high dimensional space; in the asymptotic limit, the solution converges to a lower dimensional attractor.

The chaotic properties of the attractor are characterized by invariant measures, such as Lyapunov exponents, the Kaplan-Yorke dimension, and the geometry of the tangent space. Although it is possible to infer these measures through echo state networks (ESNs) \cite{Margazoglou2023stability} and long short-term memory networks \cite{ozalp2023reconstruction}, these methods for data-driven stability analysis do not effectively apply to high-dimensional data.  Furthermore, the stability properties of hybrid CAE-RNN models, specifically of the latent manifolds generated by autoencoders, remain unexplored.

In this paper, we address this gap by employing ESNs, a reservoir computer, as recurrent neural networks \cite{jaeger2001echo} and applying stability analysis to investigate and characterize latent manifolds.  We thus extend the CAE-ESN methodology \cite{racca2023predicting} by performing data-driven stability analysis on the latent space of a convolutional autoencoder of a spatiotemporal PDE, i.e., the Kuramoto-Sivashinsky system.

\section{Learning stability properties in latent spaces}
The Kuramoto-Sivashinsky (KS) equation is a prototypical partial differential equation (PDE) to study chaos in spatial-dissipative systems \cite{Kuramoto_1978_DiffusionInducedChaos}. The PDE models instabilities and flame fronts and is given by
\begin{equation}\label{eq:ks_equation}
\mathbf{u}_t+ \mathbf{u}_{\mathbf{x}\mathbf{x}}+ \mathbf{u}_{\mathbf{x}\mathbf{x}\mathbf{x}\mathbf{x}}+ \mathbf{u}\mathbf{u}_{\mathbf{x}}  = 0
\end{equation} 
with periodic boundary conditions on the spatial domain $[0, L]$. To solve the PDE, we spatially discretize the domain into $N_x$ grid points. This discretization allows us to interpret Eq.~\eqref{eq:ks_equation} as a dynamical system in the form of $\mathbf{u}_t = f(\mathbf{u})$ with $f$ being a smooth function and $\mathbf{u} \in \mathbb{R}^{N_x}$. 

As the domain length $L$ increases, the solution of Eq.~\eqref{eq:ks_equation} transitions from a quasiperiodic system to a chaotic solution \cite{papageorgiou1991route}. This behaviour can be analysed via stability analysis, which focuses on the tangent space defined by the Jacobian $J = {\partial f}/{\partial \mathbf{u}}$ to examine the chaotic stability. By imposing an infinitesimal perturbation $ \boldsymbol{\delta} \mathbf{u}$ to the trajectory of the system and evolving it after Eq.~\eqref{eq:ks_equation}, we obtain the linearized tangent equation
\begin{equation}\label{tangent_eq}
    \boldsymbol{\delta} \mathbf{u}_t = J(\mathbf{u}(t)) \boldsymbol{\delta} \mathbf{u}.
\end{equation}
For a detailed explanation of tangent space methods, we refer the reader to \cite{Ginelli2007_CharacterizingDynCLVs, Sandri_1996_NumericalCalculationLyapunovExponents} and briefly summarize the key properties that can be derived from Eq.~\eqref{tangent_eq} here.

Chaotic systems are characterized by exponential sensitivity to perturbations, a property which can be measured through Lyapunov exponents (LEs). Each Lyapunov exponent $\lambda_1, \dots, \lambda_{N_x}$ measures the average rates of convergence and divergence of nearby trajectories in the system's phase space $\mathbb{R}^{N_x}$. When the system is chaotic, the solution $\mathbf{u}$ occupies only a small subset $\mathcal{A}$ of the phase space in $\mathbb{R}^{N_x}$. This subset, known as the attractor, has a fractal structure, the dimension of which can be estimated through the Kaplan-Yorke dimension $D_{KY}$\cite{frederickson1983liapunov}, among other dimensionality estimatess \cite{strogatz2018nonlinear, kuptsov2012theory}.

Further insight into the system’s geometry is provided by the covariant Lyapunov vectors (CLVs) $\mathbf{v}_1, \dots, \mathbf{v}_{N_x}$ \cite{Ginelli2007_CharacterizingDynCLVs,huhn2020stability}, which offer an invariant splitting of the tangent space into unstable, neutral, and stable subspaces, corresponding to positive, zero, and negative Lyapunov exponents, respectively. In hyperbolic systems, the angles between CLVs from different subspaces are bounded away from zero. The angle $\theta_{\mathbf{v}_i, \mathbf{v}_j}$ between a pair of CLVs $\mathbf{v}_i$ and $\mathbf{v}_j$ from different subspaces is calculated as
\begin{equation}
\theta_{\mathbf{v}_i, \mathbf{v}_j} = \frac{180^{\circ}}{\pi} \cos^{-1}(|\mathbf{v}_i \cdot \mathbf{v}_j|),
\end{equation}
where $\cdot$ denotes the dot product. For spatial-dissipative systems such as the KS system, the CLV angles also allow for the isolation of physically relevant modes that capture the essential dynamics of the system. The number of these physically relevant modes is typically larger than the Kaplan-Yorke dimension but still significantly smaller than the number of grid points. From a theoretical perspective, it has been conjectured that capturing the physical modes is sufficient to faithfully integrate PDEs \cite{takeuchi2011hyperbolic}. This provides a guideline for the design of an optimal reduced order model,  which should capture all the time-invariant properties of the attractors, i.e. the LEs and the Kaplan-Yorke dimension, but also reproduce the geometric structure of the attractor through the CLVs. 

\section{Methods}
For dissipative chaotic systems such as in Eq.~\eqref{eq:ks_equation}, the state space may be high-dimensional, but in the asymptotic limit, the solution converges to a lower-dimensional chaotic attractor. This work follows a two-step approach: first, we compute the low-dimensional manifold with a convolutional autoencoder (CAE) for a nonlinear reduced-order representation, and second, we propagate the latent dynamics using an echo state network (ESN), see Fig.~\ref{fig:ks_autoencoder_potatoe_esn}.\footnote{All code is available on GitHub: \href{https://github.com/MagriLab/LatentStability}{github.com/MagriLab/LatentStability}. The training is performed on a single Quadro RTX 8000.}
\subsection{Convolutional autoencoder}
Let $\{ \mathbf{u}(t_i, \mathbf{x})\}_{i=0, 1, \dots}$  be the discretized solution to Eq.~\eqref{eq:ks_equation}.  We first aim to find a lower-dimensional manifold representation of the system using a CAE. The encoder $\mathcal{E}$ maps the physical input $\mathbf{u}(t_i) \in \mathbb{R}^{N_{x}}$ to a latent representation $ \mathbf{y}(t_i) \in \mathbb{R}^{N_{lat}}$. The decoder $\mathcal{D}$ then maps the physical state from the latent representation to the phase space 
\begin{align}
    \mathbf{\hat{u}}(t_i) \approx \mathbf{u}(t_i), \quad \text{where } \mathbf{\hat{u}}(t_i) = \mathcal{D}\left( \mathbf{y}(t_i) \right), \quad \mathbf{y}(t_i)=\mathcal{E}\left(\mathbf{u}(t_i) \right),
\end{align}
and $\mathbf{\hat{u}}(t_i)$ is the autoencoder reconstruction. The network is trained by minimizing the loss $ \mathcal{L} (\mathbf{u}, \mathbf{\hat{u}}) = \frac{1}{N_{tr}} \sum_{i=1}^{N_{tr}} \| \mathbf{u}(t_i) - \mathbf{\hat{u}}(t_i) \|_2^2, $ where $N_{tr}$ is the number of training samples. By first applying an autoencoder to each snapshot, we obtain a spatial compression of the input, with the convolutional layers filtering the spatial multiscale structure of the data. The objective is to find a latent space that approximately retains the geometry of the attractor while providing a minimal representation.

\subsection{Echo state network }
The trained autoencoder provides a transformation from the physical snapshot $\mathbf{u}(t_i)$ to the latent representation $\mathbf{z}(t_i)$ and vice versa. To learn the temporal dynamics of the training data, we employ the ESN \cite{jaeger2001echo} on the time-ordered latent representations $\{ \mathbf{y}(t_i)\}_{i=1, \cdots, N_{tr}}$, resulting in the CAE-ESN. The ESN evolves the latent representations $\mathbf{y}(t_i)$ to the next-time step prediction $\mathbf{\Hat{y}}(t_{i+1})$ according to
\begin{align}\label{eq:esn1}
        \mathbf{r}(t_{i+1}) &= \tanh\left([\mathbf{y}(t_i); 1]^T\mathbf{W}_{in} + \mathbf{r}(t_i)^T\mathbf{W} \right), \\
         \mathbf{\Hat{y}}(t_{i+1}) & = [\mathbf{r}(t_{i+1}), 1]^T \mathbf{W}_{out}.
\end{align}
Here, the matrices $\mathbf{W}_{in}$ and $\mathbf{W}$ are pseudorandomly generated and fixed \cite{lukovsevivcius2012practical} and $\sigma_{in}$ is found by Bayesian optimization \cite{racca2021robust}. By formulating the loss as $\mathcal{L}(\mathbf{y},\mathbf{\hat{y}}) =\frac{1}{N_{tr}} \sum_{i=1}^{N_{tr}} \| \mathbf{y}(t_i)- \mathbf{\Hat{y}}(t_{i})\|_2^2$, the ESN training is performed by solving for $\mathbf{W}_{out}$ using ridge regression, speeding up the training by avoiding backpropagation. The ESN can be employed in two modes, open-loop and closed-loop. Whilst training, the network operates in open-loop mode, predicting the next time step based on a reference input. Once $\mathbf{W}_{out}$ has been fixed, the network can be employed in the closed-loop configuration where the network prediction is used as input in Eq.~\eqref{eq:esn1}, allowing the network to autonomously evolve without additional input data. 

This setup effectively defines a dynamical system, and the stability properties of the network can be calculated \cite{Margazoglou2023stability}. The first step is to calculate the Jacobian of the ESN,  $ \mathbf{J}_{esn}(\mathbf{r}(t_{i+1})) = (1-\mathbf{r}(t_i)^2) ( \mathbf{W_{in}^T \mathbf{W}_{out}^T} + \mathbf{W}^T)$. The Jacobian $\mathbf{J}_{esn}$ is then employed in the tangent equation, Eq.~\eqref{tangent_eq}, to calculate the LEs and CLVs \cite{Margazoglou2023stability}.
The ESN predicts the temporal dynamics on the latent manifold and the Jacobian, therefore, defines the tangent space of the latent manifold. Consequently, the calculated stability properties are inferred on the latent manifold, not in the full physical space.
\section{Numerical tests}
We consider the KS equation with $L=22$ for which the system is chaotic with an attractor dimension of $D_{KY}=6.007$ and a largest Lyapunov exponent of $0.045$ \cite{gupta2023mori}. In related work \cite{cvitanovic2010state, linot2020deep}, the number of physical modes is determined to be $8$, providing an estimate for the minimal autoencoder latent dimension. We generate the reference solution of the Kuramoto-Sivashinsky equation with $N_x = 512$ for a duration of $T=10^5$ and $\Delta t = 0.05$. 

\begin{figure}[h]
    \centering
    \includegraphics[width=\linewidth]{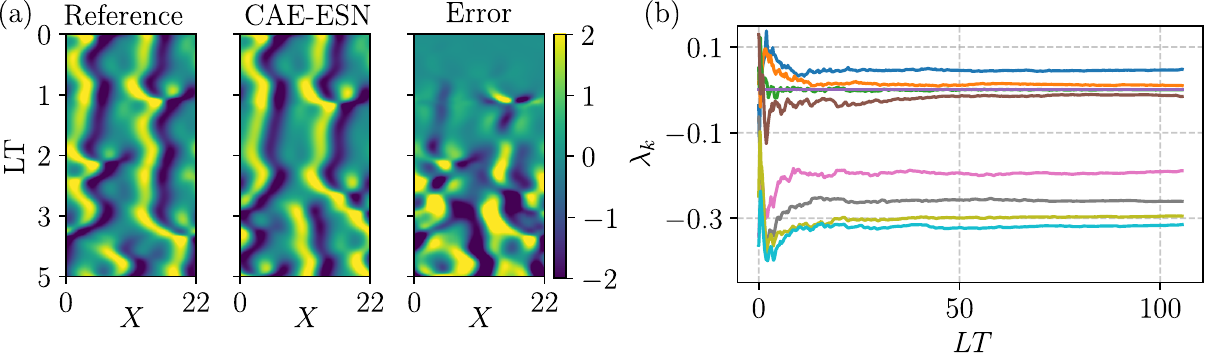}
    \caption{(a) Autonomous prediction of the CAE-ESN on the latent manifold of dimension $N_{lat}=8$. (b) Convergence of the Lyapunov exponents of the CAE-ESN over $100LT$.}
    \label{fig:pred_convergence}
\end{figure}

First, we train a convolutional autoencoder to map the full state $\mathbf{u}(t_i)$ with $N_x = 512$ to the latent representation $\mathbf{z}(t_i)$ with dimension $N_{lat}=8$. After training the autoencoder, we train the ESN on the latent space. In Fig.~\ref{fig:pred_convergence}(a), we present the autonomous prediction of the CAE-ESN based on the reduced representation. Despite using less than $2\%$ of degrees of freedom, the CAE-ESN accurately forecasts the evolution for about $2LT$ \footnote{One Lyapupunov time (LT) is defined by $\lambda_1^{-1}$ and denotes a characteristic time scale at which two trajectories of a chaotic system diverge. }. After $2LT$ the trajectories slowly diverge, reflecting the inherent sensitivity of chaotic systems to perturbations.

\begin{figure}[h]
  \centering
  \includegraphics[width=\linewidth]{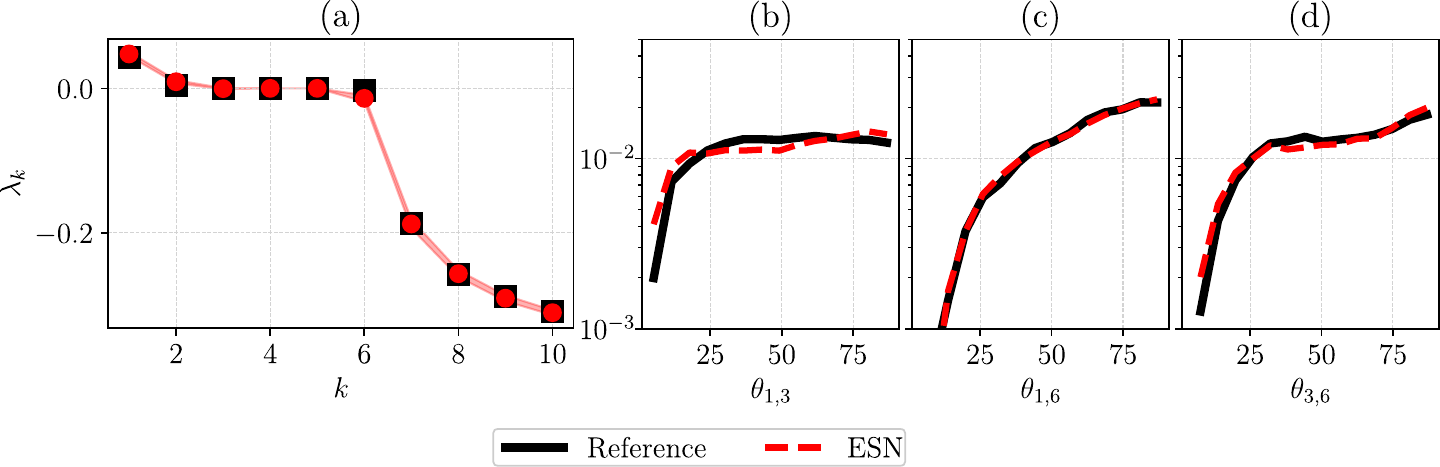}
\caption{(a) The first $10$ Lyapunov exponents of the dynamical system (black squares) compared to the Lyapunov exponents across 10 different ESNs (red dots) after $100LT$. The shaded area indicates the standard deviation. (b - d) The angle distribution of the Kuramoto–Sivashinsky system for leading covariant Lyapunov vectors of the (b) unstable-neutral, (c) unstable-stable and (d) neutral-stable Lyapunov exponents.}
\label{fig:KS_LE_CLV_L22}
\end{figure}

 More insight into the physical behaviour of the CAE-ESN is obtained by investigating the invariant properties of the network.  In Fig.~\ref{fig:pred_convergence}(b), we show the convergence of the LEs of the CAE-ESN over $100LT$. To assess the robustness, we take an ensemble of $10$ ESNs to analyse the stability properties in the latent space. In Fig.~\ref{fig:KS_LE_CLV_L22}(a), we compare the reference LEs (in black squares) with the mean LEs obtained from $10$ ESNs (in red circles). The mean Kaplan-Yorke Dimension of the CAE-ESN is given by $6.008 \pm 5\cdot10^4$ compared to the reference of $D_{KY}=6.007$. The CAE-ESNs correctly reproduce the first $10$ LEs, which correspond to the physical behaviour of the system, indicating that the forecast based on the reduced representation captures all essential chaotic dynamics. 

We further examine the CAE-ESN abilities to capture the geometric structure by studying the angles between CLVs in  Fig.~\ref{fig:KS_LE_CLV_L22}(b - d). The agreement of angle statistics between the reference and the CAE-ESN is within a negligible numerical error with a Wasserstein distance of $0.001$, demonstrating that the geometrical structure of the attractor has been effectively captured in the latent manifold. 
\section{Conclusions}
We propose a data-driven method to infer the stability properties of chaotic partial differential equations. This is based on the convolutional autoencoder echo state network (CAE-ESN), where the CAE discovers a nonlinear latent representation of minimal dimension, on which the ESN propagates the temporal dynamics. By applying the CAE-ESN to the chaotic Kuramoto-Sivashinsky equation, we show that (i) invariant measures, such as Lyapunov exponents and the Kaplan-Yorke dimension, are accurately and robustly inferred in the latent space,  (ii) the geometric properties of the attractor, including the angles of the covariant Lyapunov vectors (CLVs), are accurately learned.
This approach enables both nonlinear forecasting and stability analysis of chaotic partial differential equations. Inferring stability properties of latent manifolds generated by a CAE offers a perspective on understanding autoencoder manifolds using dynamical systems tools. 
\printbibliography


\end{document}